\documentclass[12pt, a4paper]{article}

\usepackage[left=25mm, right=25mm, top=25mm, bottom=25mm]{geometry}
\usepackage{mathptmx}
\usepackage[T1]{fontenc}
\usepackage{setspace}
\usepackage{titlesec}
\titleformat{\section}[block]
  {\normalfont\fontsize{16}{19}\bfseries\centering}
  {}{0em}{\MakeUppercase}

\titleformat{\subsection}[block]
  {\normalfont\fontsize{14}{17}\bfseries\flushleft}
  {}{0em}{\MakeUppercase}

\titleformat{\subsubsection}[block]
  {\normalfont\fontsize{12}{14}\bfseries\flushleft}
  {}{0em}{}

\titleformat{\paragraph}[runin]
  {\normalfont\fontsize{12}{14}\bfseries\itshape}
  {}{0em}{}

\usepackage[numbers]{natbib}
\setcitestyle{open={(},close={)}}

\usepackage{graphicx}
\usepackage{amsmath}
\usepackage{caption}
\usepackage{amssymb}
\usepackage[mathcal]{euscript}
\usepackage{xurl}
\usepackage{cleveref}
\usepackage{subcaption}
\usepackage[utf8]{inputenc}
\usepackage{booktabs}  % 核心套件：畫漂亮的橫線
\usepackage{makecell}  % 核心套件：支援單元格內換行
\usepackage{multirow}
\usepackage{enumitem}
\usepackage{array}
\usepackage{IEEEtrantools}

\usepackage[cal=cm, scr=rsfs]{mathalpha}

\titlespacing*{\section}{0pt}{6pt plus 2pt minus 2pt}{6pt plus 2pt minus 2pt}
\titlespacing*{\subsection}{0pt}{4pt plus 2pt minus 2pt}{4pt plus 2pt minus 2pt}
\titlespacing*{\subsubsection}{0pt}{1pt plus 1pt minus 1pt}{1pt plus 1pt minus 1pt}

\begin{document}
\bstctlcite{IEEEexample:BSTcontrol}

% --- 標題頁 ---
\begin{center}
    % \vspace*{1em}
    {\fontsize{18}{22}\selectfont\bfseries\uppercase{A Robust and Efficient Multi-Agent Reinforcement Learning Framework for Traffic Signal Control}}
    \vspace{2em}
    
    % {\fontsize{12}{14}\bfseries
    % Author Name 1\\
    % Position, Organization\\
    % Mailing Address\\
    % TEL, E-mail
    
    % \vspace{0.5em}
    
    % Author Name 2\\
    % Position, Organization\\
    % Mailing Address\\
    % TEL, E-mail
    
    % \vspace{0.5em}
    
    % Author Name 2\\
    % Position, Organization\\
    % Mailing Address\\
    % TEL, E-mail
    
    % }

  %   \begin{tabular}{ccc} 
  %   % --- 第一排 (Authors 1-3) ---
  %   \textbf{Sheng-You, Huang} & \textbf{Hsiao-Chuan, Chang} & \textbf{Yen-Chi, Chen} \\
  %   National Yang Ming Chiao Tung University & National Yang Ming Chiao Tung University & Academia Sinica \\
  %   \texttt{shengyouhuang620@gmail.com} & \texttt{zekeabc@gmail.com} & \texttt{zxkyjimmy@gmail.com} \\[3ex] % [3ex] 是行距，可自行調整
    
  %   % --- 第二排 (Authors 4-6) ---
  %   \textbf{Ting-Han, Wei} & \textbf{Author Name 5} & \textbf{Author Name 6} \\
  %   Kochi University of Technology & Organization 5 & Organization 6 \\
  %   \texttt{tinghan.wei@kochi-tech.ac.jp} & \texttt{email5@domain.com} & \texttt{email6@domain.com} \\[3ex]
    
  %   % --- 第三排 (Authors 7-9) ---
  %   \textbf{Author Name 7} & \textbf{Author Name 8} & \textbf{I-Chen, Wu} \\
  %   Organization 7 & Organization 8 & National Yang Ming Chiao Tung University \\
  %   \texttt{email7@domain.com} & \texttt{email8@domain.com} & \texttt{icwu@cs.nycu.edu.tw}
  % \end{tabular}
    \vspace{-0.5em}
    \fontsize{12}{14}\selectfont 
    \begin{tabular}{cc} 
    % --- 第一排 (Authors 1-3) ---
    
    \textbf{Sheng-You Huang\textsuperscript{*}} & \textbf{Hsiao-Chuan Chang\textsuperscript{*}} \\
    National Yang Ming Chiao Tung University & National Yang Ming Chiao Tung University \\
    \texttt{shengyouhuang.cs13@nycu.edu.tw} & \texttt{s312581015.ii12@nycu.edu.tw} \\[1ex] % [3ex] 是行距，可自行調整

    \textbf{Yen-Chi Chen} & \textbf{Ting-Han Wei} \\
    Academia Sinica & Kochi University of Technology \\
    \texttt{zxkyjimmy@gmail.com} & \texttt{tinghan.wei@kochi-tech.ac.jp} \\[1ex]
    
    % --- 第二排 (Authors 4-6) ---
    \textbf{I-Hau Yeh} & \textbf{Sheng-Yao Kuan} \\
    ELAN Microelectronics Corporation & ELAN Microelectronics Corporation \\
    \texttt{ihyeh@emc.com.tw} & \texttt{shaun.kuan@emc.com.tw} \\[1ex]
    
    % --- 第三排 (Authors 7-9) ---
    \textbf{Chien-Yao Wang} & \textbf{Hsuan-Han Lee} \\
    Academia Sinica & ELAN Microelectronics Corporation \\
    \texttt{kinyiu@iis.sinica.edu.tw} & \texttt{henry.lee@emc.com.tw} \\[1ex]

    \textbf{I-Chen Wu} \\
    National Yang Ming Chiao Tung University \\
    \texttt{icwu@cs.nycu.edu.tw} \\[1.3ex]

    \multicolumn{2}{c}{
        \fontsize{11}{13}\selectfont 
        \textit{* Equal Contribution.}
    }
  \end{tabular}
\end{center}

\vspace{-0.7em}

% --- Abstract ---
\section*{ABSTRACT}
% Reinforcement Learning (RL) has demonstrated significant potential for Traffic Signal Control (TSC); however, bridging the gap between simulation and real-world deployment remains a challenge due to poor environmental generalization and safety concerns regarding erratic signal switching. Existing RL approaches often overfit to static traffic patterns and utilize action spaces that violate driver expectations. 
% This paper proposes a robust and stable Multi-Agent Reinforcement Learning (MARL) framework validated on a high-fidelity Vissim simulation network. 
% We introduce three key innovations to address current limitations: 
% (1) a Turning Ratio Randomization training strategy that prevents overfitting by exposing agents to dynamic turning probabilities, thereby enhancing robustness against unseen traffic scenarios; 
% (2) Stability-Oriented Exponential Action Space that adjusts phase duration cyclically using exponential increments, striking a balance between rapid responsiveness and fine-grained precision; 
% and (3) an optimized Neighbor-based Observation scheme leveraging Centralized Training Decentralized Execution (CTDE) to approximate global coordination with scalable local communication. 
% Experimental results demonstrate that our framework significantly outperforms fixed-time control and standard RL baselines. 
% Notably, the proposed model exhibits superior generalization capabilities across varying traffic volumes (ranging from 0.7x to 1.0x) and maintains high control stability, offering a practical and safe solution for adaptive traffic signal control.

Reinforcement Learning (RL) in Traffic Signal Control (TSC) faces significant hurdles in real-world deployment due to limited generalization to dynamic traffic flow variations. 
Existing approaches often overfit static patterns and use action spaces incompatible with driver expectations. 
This paper proposes a robust Multi-Agent Reinforcement Learning (MARL) framework validated in the Vissim traffic simulator. 
The framework integrates three mechanisms:
(1) Turning Ratio Randomization, a training strategy that exposes agents to dynamic turning probabilities to enhance robustness against unseen scenarios; 
(2) a stability-oriented Exponential Phase Duration Adjustment action space, which balances responsiveness and precision through cyclical, exponential phase adjustments; 
and (3) a Neighbor-Based Observation scheme utilizing the MAPPO algorithm with Centralized Training with Decentralized Execution (CTDE). 
By leveraging centralized updates, this approach approximates the efficacy of global observations while maintaining scalable local communication.
Experimental results demonstrate that our framework outperforms standard RL baselines, reducing average waiting time by over 10\%. 
The proposed model exhibits superior generalization in unseen traffic scenarios and maintains high control stability, offering a practical solution for adaptive signal control.

\textbf{Key word:} Multi-agent RL, TSC optimization, CTDE

% --- Main Text ---

\section{Introduction}
As urbanization accelerates globally, traffic congestion has evolved into a critical bottleneck restricting sustainable city development, causing severe economic losses and environmental degradation. 
According to the INRIX traffic report in 2025 \cite{inrix_scorecard_}, traffic congestion costs the U.S. economy over \$85 billion—an 11.3\% increase from the previous year. 
Drivers in metropolitan areas continue to lose approximately $50 \sim 112$ hours annually to delays, underscoring the critical need for Advanced Traffic Management Systems (ATMS) to mitigate these severe economic and temporal losses.
Traditional control methods rely on predefined rules and often lack the flexibility to handle the non-linear and stochastic nature of real-time traffic. 
Conversely, Deep Reinforcement Learning (DRL) enables agents to dynamically optimize signal policies, offering superior adaptability to fluctuating traffic demands.

Direct training in physical environments is prohibited by safety and cost constraints, making high-fidelity simulation indispensable. 
While most Reinforcement Learning Traffic Signal Control (RL-TSC) studies use SUMO \cite{krajzewicz_sumo_2002} or CityFlow \cite{zhang_cityflow_2019} for their efficiency, these platforms often rely on simplified models that lack realism.  
To bridge the 'sim-to-real' gap, this study uses PTV Vissim \cite{_traffic_}, the industry standard for microscopic traffic modeling. 
Addressing the scarcity of Vissim-based RL research due to complex integration barriers, we leverage the VissimRL framework \cite{Chang_VissimRL_IV} to seamlessly interface the RL agent with this high-fidelity environment.

Despite advanced simulation tools, real-world deployment faces three fundamental hurdles. 
First, regarding environmental generalization, agents must adapt to stochastic, non-stationary traffic flows without overfitting to static training patterns. 
Second, action space design requires balancing stability with responsiveness—maintaining safety-critical cyclic sequences while reacting agilely to congestion. 
Finally, centralized systems do not scale to larger traffic grids. 
A framework that can facilitate global cooperation through decentralized execution is necessary.
% Finally, large-scale network control confronts the scalability-coordination dilemma, necessitating a framework that achieves global cooperation through decentralized execution to avoid the prohibitive overhead of centralized control.

Thus, we propose a robust and stable Multi-Agent RL (MARL) framework for traffic signal control (TSC), with three specific technical improvements:
\begin{enumerate}[topsep=-5pt, nosep]
    \item \textbf{Robustness via Turning Ratio Randomization:} 
        % V1
        % We introduce a randomization training strategy that dynamically perturbs intersection turning ratios. 
        % This approach enhances the agent's robustness and enables effective generalization across highly diverse and non-stationary traffic conditions.
        % V2
        We introduce a randomization training strategy to enhance the agent’s robustness across non-stationary traffic conditions.
    \item \textbf{Exponential Phase Duration Adjustment:} 
        % V1
        % We propose a cyclic control mechanism with exponential adjustment steps $\Delta t \in \{0, \pm \lambda ^0, \pm \lambda ^1, \pm \lambda ^2, \pm \lambda ^3, \dots\}$. 
        % This design facilitates 'coarse-to-fine' control, reconciling precise fine-tuning for steady-state stability with rapid responsiveness to congestion shockwaves.
        % V2
        We propose a cyclic control mechanism with exponential adjustment steps to balance stability and reactivity via coarse-to-fine control.
    \item \textbf{Scalable Coordination via Neighbor-Level Observation:} 
        % V1
        % By leveraging a centralized update mechanism to guide agents with neighbor-level observations, our approach approximates the performance of global-observation strategies while minimizing communication overhead, effectively resolving the scalability-coordination dilemma.
        % V2
        We employ a centralized training with decentralized execution (CTDE) framework to resolve the scalability and coordination dilemma through neighbor-level observations.
\end{enumerate}

% V1
% Experimental results demonstrate the efficacy of our framework.
% Specifically, the proposed Turning Ratio Randomization strategy significantly enhances robustness, reducing average waiting time by over 10\% in unseen scenarios compared to standard training methods. 
% Furthermore, the Neighbor-Based CTDE framework successfully bridges the scalability-optimality gap, enabling agents with limited neighbor-level observations to approximate the performance of global-view baselines.

% V2
Experiment results demonstrate that our framework reduces average waiting time by over 10\% in unseen scenarios, significantly enhancing robustness. 
Furthermore, our approach achieves robust coordination relying solely on scalable neighbor observations.

\section{Background}
This section establishes the theoretical foundation for our study. 
We first survey prior RL-based control methods, followed by the formalization of the TSC problem. 
Finally, we investigate the critical role of observation scope in large-scale network coordination.

\subsection{Reinforcement Learning Approaches for TSC}
Reinforcement Learning (RL) has been extensively investigated for Traffic Signal Control (TSC). 
Existing studies can be broadly categorized based on their training scenarios and action space formulations. 
However, while these approaches have achieved success in specific simulation settings, critical challenges regarding environmental generalization and control stability persist.

\subsubsection{Training Scenarios and Generalization} \label{background: overfitting}
% 2.1.1 V1
% A significant portion of existing research \cite{zang_metalight_2020, ye_fairlight_2023} focuses on optimizing performance under fixed, static traffic conditions. 
% In these setups, agents are trained and evaluated on identical traffic flow rates and turning ratios. 
% While this methodology allows agents to converge effectively to an optimal policy for specific demand profiles, it often results in models that specialize heavily in the training distribution. 
% The agents tend to memorize the optimal timing for a specific traffic demand rather than learning the underlying dynamics of traffic flow. 
% Consequently, models trained on static patterns may struggle to generalize to unseen scenarios, often necessitating the training of separate models for different time periods (e.g., distinct policies for peak versus off-peak hours) to maintain optimal performance, thereby increasing the complexity of practical implementation.
% This lack of adaptability hinders real-world deployment, where traffic conditions fluctuate continuously and unpredictably, and a single robust model is required to handle diverse scenarios without frequent retraining.

% 2.1.1 V2
A substantial body of research \cite{zang_metalight_2020, ye_fairlight_2023} focuses on optimizing performance under static traffic conditions, where flow rates and turning ratios remain constant. 
Although this setup ensures stable convergence for specific demand profiles, it often leads to overfitting the training distribution. 
Instead of learning the underlying traffic dynamics, agents tend to memorize specific timing patterns. 
Consequently, these models struggle to generalize to unseen scenarios, often requiring distinct policies for different time periods (e.g., peak vs. off-peak) to maintain performance. 
This limitation hinders real-world deployment, where traffic is non-stationary, necessitating a single robust model capable of handling diverse conditions without frequent retraining.

\subsubsection{Action Space Architectures} \label{background: action space design}
Action space design is critical for ensuring both the safety and stability of control policies. 
Existing approaches typically adopt either acyclic or cyclic control mechanisms.

Acyclic control \cite{ye_fairlight_2023, wei_presslight_2019, wei_colight_2019, oroojlooy_attendlight_2020, chen_thousand_2020} offers flexibility by allowing arbitrary phase selection but violates the fixed-sequence mandates essential for driver safety.
Consequently, cyclic control is preferred for real-world deployment.
Binary Switching methods \cite{cabrejas-egea_reinforcement_2021, kumarasamy_integration_2024, zhou_drle_2021} provides fine-grained control but often leads to unpredictable signal oscillation and high communication overhead.
Direct Duration Setting methods \cite{zhang_phase_2024, park_integrated_2024}, although simple, frequently induces high temporal variance between cycles.
Finally, Duration Adjustment methods \cite{wang_traffic_2024, shah_deep_2024} enhance stability by incrementally updating durations, yet they typically rely on fixed linear steps (e.g., $\Delta t \in \{0, \pm 3, \pm 6\}$).
This fixed granularity creates an inherent trade-off: linear steps are often too small to counter sudden congestion surges or too large to maintain steady-state stability.

\subsection{Problem Formulation} \label{background: problem formulation}
\begin{figure}[h] % [h] 代表盡量放在當前位置 (here)
    \centering % 讓圖片置中
    \vspace{-0.5em}
    \includegraphics[width=0.7\textwidth]{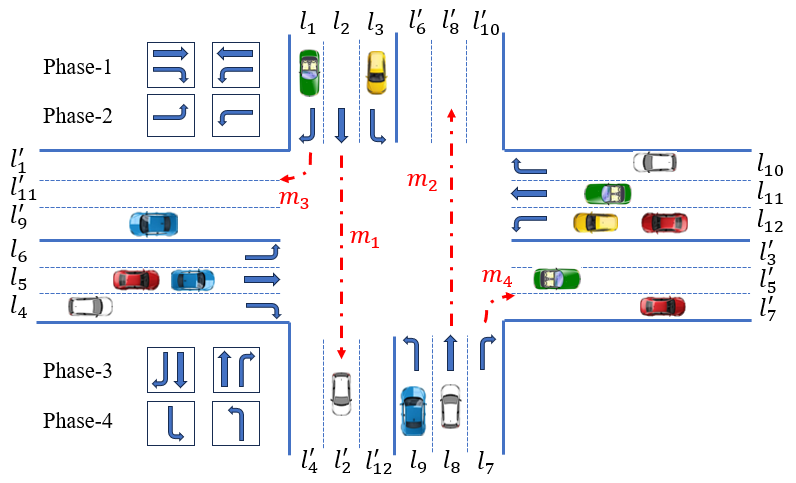} % 設定寬度為頁面文字寬度的 80%
    \vspace{-0.5em}
    \caption{Traffic Terminology} % 圖片標題
    \label{fig:Traffic terminology} % 用於參照的標籤
    \vspace{-0.5em}
\end{figure}

We first define the fundamental structural components of an intersection (shown in \Cref{fig:Traffic terminology}):
\begin{itemize}[topsep=-5pt, nosep]
    \item Incoming and outgoing Lanes ($l_i$ and $l_i^{'}$ in \Cref{fig:Traffic terminology}): 
        The set of lanes carrying traffic approaching or departing from  the intersection. 
        % Incoming lanes ($l_i$ in \Cref{fig:Traffic terminology}) are the sources of traffic demand and queue, and outgoing lanes ($l_i^{'}$ in \Cref{fig:Traffic terminology}) receive the traffic discharged.
    \item Movements ($m$): 
        A specific traffic stream transitioning from a subset of incoming lanes to a subset of outgoing lanes. For instance, movement $m_1$ includes $l_2$ in \Cref{fig:Traffic terminology}.  Let $\mathcal{M}$ denote the set of all valid movements at an intersection.
    \item Phases ($p$): 
        A set of non-conflicting movements that are allowed to proceed simultaneously.
        For instance, $\text{Phase-3}$ includes $\{m_1, m_2, m_3, m_4\}$ in \Cref{fig:Traffic terminology}.
\end{itemize}

We model the traffic signal control problem across a network of $N$ intersections as a \textbf{Decentralized Partially Observable Markov Decision Process (Dec-POMDP)}.
Each intersection operates as an autonomous agent that interacts with the environment. 
Formally, the Dec-POMDP is defined by the tuple $\langle \mathcal{S}, \mathcal{A}, \mathcal{P}, \mathcal{R}, \mathcal{O}, \gamma \rangle$, where:
\begin{itemize}[topsep=-5pt, nosep]
    \item $\mathcal{S}$: 
        Denotes the global state space of the environment, including real-time traffic distributions and the current signal control status. 
    \item $\mathcal{O}$:
        Represents the set of joint observations. 
        Each agent $i$ receives a local observation $o_{i,t} \in \mathcal{O}$ derived from the global state $\mathbf{s}_t$ according to an observation function. 
    \item $\mathcal{A}$: 
        Represents the set of joint action space.  
        At each time step $t$, agents execute a joint action $\mathbf{a}_t = \{a_{1,t}, \dots, a_{N,t}\} \in \mathcal{A}$. 
    \item $\mathcal{P}(\mathbf{s}_{t+1} | \mathbf{s}_t, \mathbf{a}_t)$: 
        Represents the state transition probability function, denoting the probability of the environment transitioning to state $\mathbf{s}_{t+1}$ by the current state $\mathbf{s}_t$ and action $\mathbf{a}_t$. 
    \item $\mathcal{R}$: 
        Is the reward function that evaluates the immediate quality of $\mathbf{a}_t$, in the form of $\mathbf{r}_t=\{r_{1,t},...,r_{N,t}\}$.
        Common reward designs include throughput or waiting time.
    \item $\gamma$: 
        $\gamma \in [0, 1)$ is the discount factor, which determines the importance of future rewards. 
\end{itemize}
The objective is to maximize the expected return $J(\pi) = \mathbb{E}[\sum_{i=1}^{N}\sum_{k=0}^{\infty} \gamma^k r_{i,t+k}]$, where $\pi = \{\pi_1, ..., \pi_N\}$ and $\pi_i(o_{i,t}) = a_{i,t}$ for the state $\mathbf{s}_t$ and agent $i$.
In this paper, we define the state $\mathbf{s}_t$ as the concatenation of the current phase ID, the time elapsed in this current phase, and a vector of vehicle counts for each lane within a limited range on approaching links, and the reward is computed using a weighted-sum of travel time (s), waiting time (s), average speed (s), and throughput (vehs).

\subsection{Observation Scopes Analysis} \label{observation scopes definition}
\begin{figure}[h] % [h] 代表盡量放在當前位置 (here)
    \centering % 讓圖片置中
    \vspace{-0.5em}
    \includegraphics[width=0.7\textwidth]{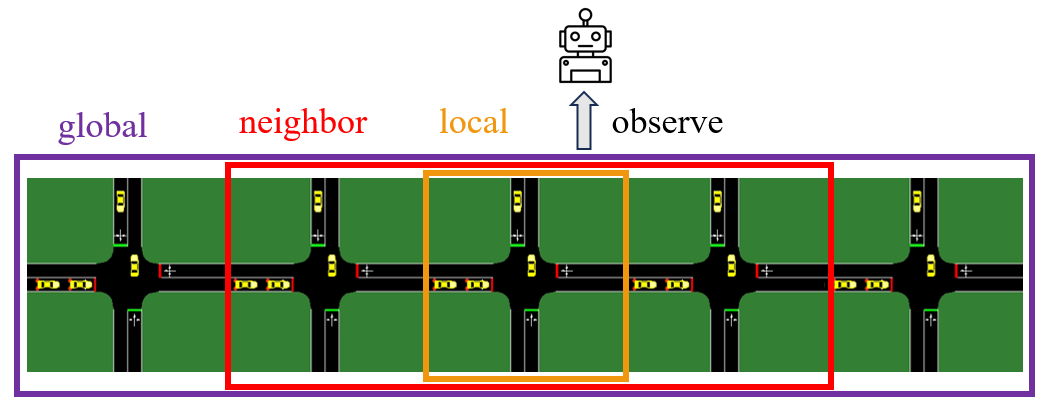} % 設定寬度為頁面文字寬度的 80%
    \vspace{-0.5em}
    \caption{Observation scope} % 圖片標題
    \label{fig:observation scope} % 用於參照的標籤
    \vspace{-1em}
\end{figure}

The scope of information available to an agent fundamentally dictates the balance between control efficacy and system scalability. 
As illustrated in \Cref{fig:observation scope} (where the agent controls the central intersection 3), we categorize observation scopes into three distinct levels.

First, local observation ($O^{\text{local}}$) restricts visibility to the host intersection. 
While efficient, this myopia prevents agents from anticipating upstream platoons, hindering green wave formation \cite{han_wavelearner_2022}.
Conversely, global observation ($O^{\text{global}}$) covers the entire network. 
Although theoretically optimal, it faces severe scalability issues; the input dimension grows linearly with network size $|N|$, yielding an unmanageable complexity of $O(|N| \times |O^{\text{local}}|)$.
% In addition, in large-scale road networks (more than 20 intersections), data transmission quality can be affected by network stability.
To resolve this, neighbor observation ($O^{\text{neighbor}}$) includes only directly connected intersections $\mathcal{N}_i$. 
This captures essential incoming flow dynamics for coordination while maintaining a constant state space size independent of the total network dimension.

\section{Methodology}
In this section, we detail the proposed MARL framework designed for real-world traffic signal control deployment. 
Our methodology focuses on three critical aspects: environmental robustness, control stability, and system scalability.

\begin{figure}[h] % [h] 代表盡量放在當前位置 (here)
    \centering % 讓圖片置中
    \vspace{-0.5em}
    \includegraphics[width=1\textwidth]{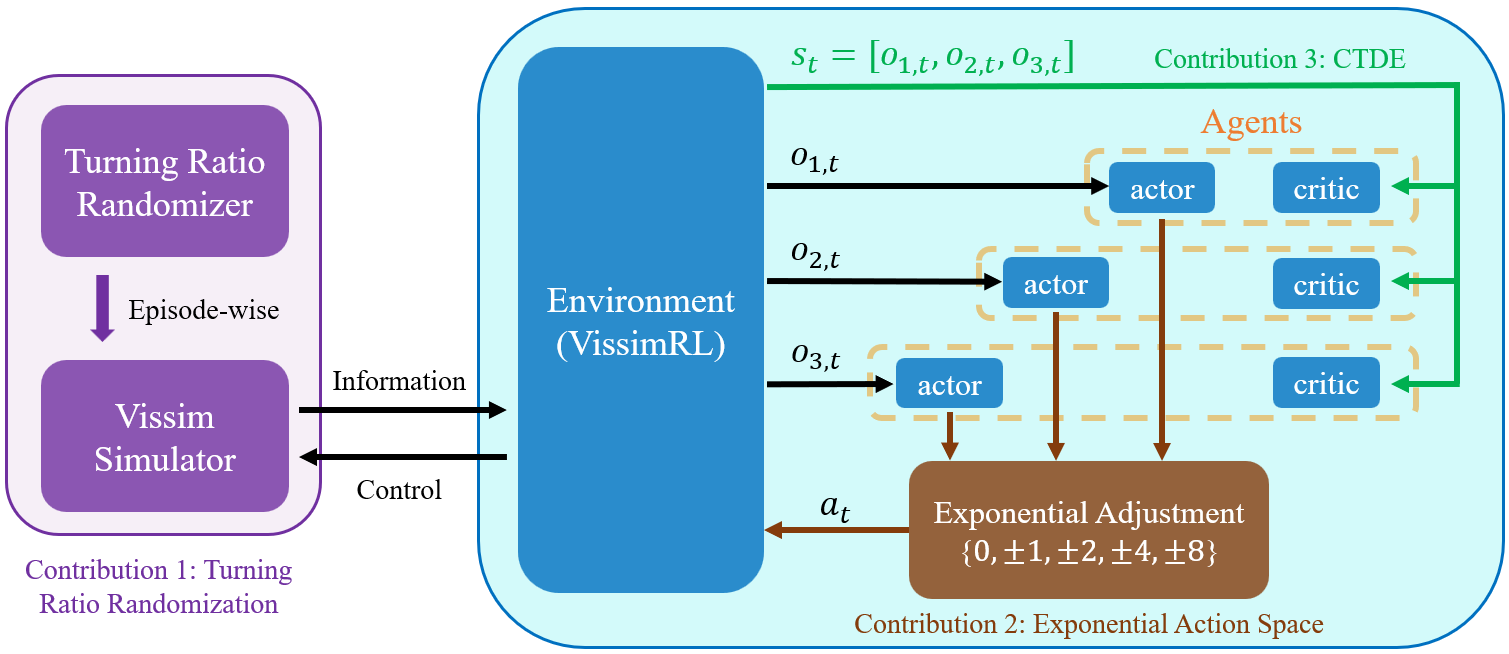} % 設定寬度為頁面文字寬度的 80%
    \caption{Method Architecture} % 圖片標題
    \label{fig: method framework} % 用於參照的標籤
    \vspace{-1em}
\end{figure}

\subsection{Turning Ratio Randomization}
Traffic dynamics are governed by two distinct factors: 
\textit{traffic volume}, which dictates the overall intersection load and required cycle length, and \textit{turning ratios}, which fundamentally determine the optimal green split—the allocation of green signal durations among competing phases. % the specific allocation of time across conflicting phases
In standard RL training environments, both traffic volume and turning ratios remain static.
Under such conditions, agents face a significant risk of overfitting. 
Instead of learning to interpret state observations, they tend to converge toward an 'open-loop' policy, implicitly memorizing a fixed timing schedule (e.g., 'switch phase after 15 seconds') that aligns with the static accumulation pattern. 
This results in a brittle policy that fails to react to real-time fluctuations.
Conversely, varying traffic volume introduces reward instability. 
Since metrics like waiting time are highly sensitive to network load, volume fluctuations cause reward magnitudes to shift independent of policy quality. 
This generates 'misleading signals' that can destabilize the learning process.

To counter this, we introduce a \textbf{Turning Ratio Randomization} strategy. 
At the beginning of each training episode, we perturb the turning probabilities of all approaches using a uniform distribution scaling method. 
Specifically, we apply independent multiplicative noise to each movement’s ratio and subsequently re-normalize the values to ensure the probabilities sum to one. 
Let $r_m$ denote the original turning ratio for movement $m$. 
The perturbed ratio $r'_m$ is computed via the following three steps:
\vspace{-0.3em}

\begin{enumerate}
    \vspace{-1em}
    \item Noise Sampling: 
        Sample a noise factor $\epsilon_m \sim U(-\delta, \delta)$ for each movement, where $\delta \in [0, 1]$ is a hyperparameter controlling the perturbation intensity.
        \vspace{-0.5em}
    \item Scaling: 
        Compute the intermediate perturbed ratio:
        \vspace{-0.3cm}
        \begin{equation}
            \vspace{-1em}
            \hat{r}_m = r_m \cdot (1 + \epsilon_m)
        \end{equation}
    \item Normalization: 
        Re-normalize to obtain the final turning ratio:
        \vspace{-0.3cm}
        \begin{equation}
            \vspace{-1em}
            r'_m = \frac{\hat{r}_m}{\sum_{k \in \mathcal{M}} \hat{r}_k}
        \end{equation}
\end{enumerate}
This approach ensures that the perturbation remains proportional to the original traffic demand, effectively preventing dominant traffic flows from being disproportionately distorted by additive noise while maintaining realistic structure.

\subsection{Exponential Phase Duration Adjustment}
To ensure safety and smooth traffic flow, we adopt a Cyclic Phase control scheme. 
Furthermore, in order to match drivers' expectations, within each phase, the order of green, yellow, red shall remain unchanged. 
Fixed durations are set for yellow and an all-phase red that is meant to clear the intersection.
Therefore, the core innovation lies in how the green duration of the next phase in the cycle is determined.
We propose an \textbf{Exponential Phase Duration Adjustment} to achieve "coarse-to-fine" control granularity.
\begin{figure}[h] % [h] 代表盡量放在當前位置 (here)
    \centering % 讓圖片置中
    \vspace{-0.5em}
    \includegraphics[width=0.9\textwidth]{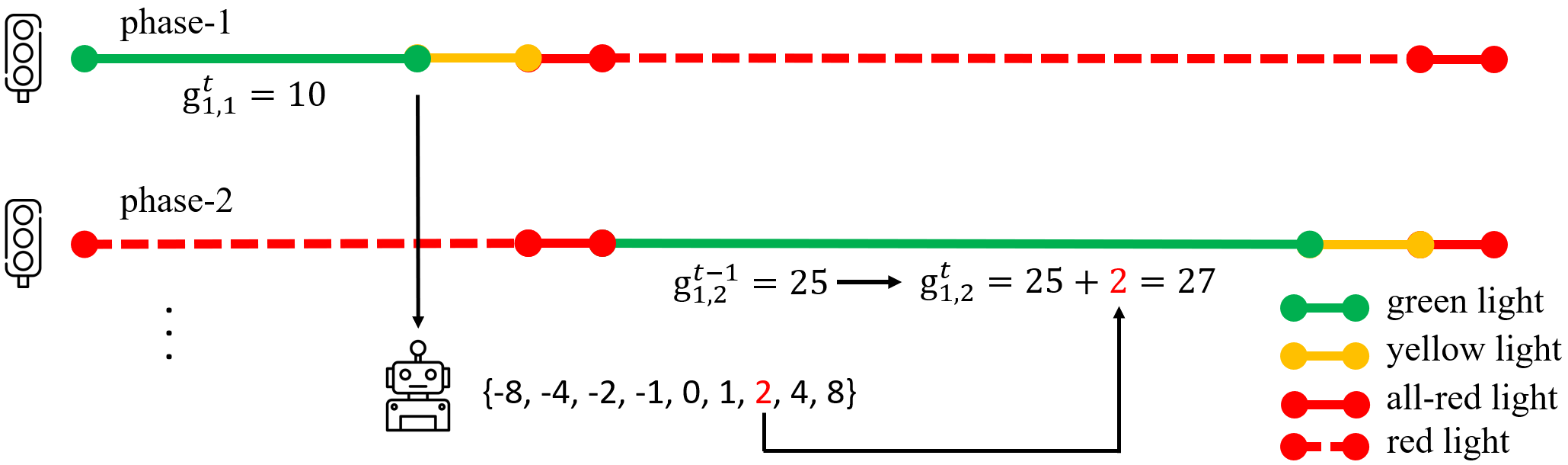} % 設定寬度為頁面文字寬度的 80%
    \caption{Action space: Set Next Phase Duration with small change} % 圖片標題
    \label{fig:Set Next Phase Duration with Small Change} % 用於參照的標籤
    \vspace{-1em}
\end{figure}

Let $g_{i, p}^{t}$ be the green light duration for phase $p$ at intersection $i$ during cycle $t$. 
The agent selects an adjustment action $\Delta t$ from a discrete exponential adjustment set:
\vspace{-0.3em}
\begin{equation}
    \vspace{-0.5em}
    \Delta t \in \{0, \pm \lambda^0, \pm \lambda^1, \pm \lambda^2, \pm \lambda^3 \}
\end{equation}
where $\lambda$ is a hyperparameter that determines the granularity of the adjustment.
The duration for the next cycle is determined at the end of the previous phase as:
\vspace{-0.3em}
\begin{equation}
    g_{i, p}^{t} = \text{clip}( g_{i, p}^{t-1} + \Delta t, g_{\text{min}}, g_{\text{max}} )
    \vspace{-0.3em}
\end{equation}
where $g_{i, p}^{t}$ is contrained by the minimum and maximum green times $g_{\text{min}}$ and $g_{\text{max}}$.

% This design addresses the limitations of linear adjustments by providing:
% \begin{enumerate}[topsep=-5pt, nosep]
%     \item Responsiveness: Large adjustments (e.g., $\pm 8s$ if $\lambda = 2$) allow the agent to react instantly to sudden traffic surges or shockwaves.
%     \item Precision: Small adjustments (e.g., $\pm 1s, 0s$) allow the agent to fine-tune signal timings when the traffic state is stable, minimizing unnecessary fluctuations and maintaining a smooth progression.
% \end{enumerate}
% This dual capability ensures the control policy is both agile in emergencies and stable in normal conditions.
In Linear Adjustment methods, short intervals will allow finer control, but require multiple cycles when large adjustments are needed.
In contrast, long intervals allow for rapid, flexible adjustment at the expense of finer control.
By integrating an exponential scale, we can enjoy both benefits with no downsides. 
The larger steps (e.g., $\pm8$s if $\lambda = 2$) provide the necessary agility to counteract sudden congestion, while the finer granularity (e.g., 0s or $\pm1$s) ensures precise timing during steady states. 
Consequently, the control policy maintains smooth progression in normal conditions without sacrificing the responsiveness required for critical traffic events.

\subsection{Scalable Coordination via Neighbor-Level Observation}
For large-scale network control, we must make the choice between global observation, which is optimal but unscalable, or local observation, which is scalable but myopic. 
To bridge this gap, we propose using neighbor-level observation empowered by the Centralized Training Decentralized Execution (CTDE) paradigm.

CTDE fundamentally decouples the information scope available during learning from that which is restricted during deployment. 
In the training stage, a centralized critic exploits global information, incorporating the joint states of all agents across the entire network, to accurately evaluate the network-wide impact of local actions. 
During execution, each agent functions as a decentralized actor, inferring optimal policies based solely on partial observations.
% \begin{figure}[h] % [h] 代表盡量放在當前位置 (here)
%     \centering % 讓圖片置中
%     % --- 第一個子圖 (a) ---
%     \begin{subfigure}[b]{0.45\textwidth}
%         \centering
%         % 這裡放左邊那張圖的檔名
%         \includegraphics[width=\textwidth]{img/CTDE-CT.png} 
%         \caption{Training Stage} % 這裡寫 (a) 的標題
%         \label{fig:ctde_train}         % 這裡設 (a) 的標籤
%     \end{subfigure}
%     \hfill % 這指令是用來把兩張圖撐開，讓它們並排
%     % --- 第二個子圖 (b) ---
%     \begin{subfigure}[b]{0.45\textwidth}
%         \centering
%         % 這裡放右邊那張圖的檔名
%         \includegraphics[width=\textwidth]{img/CTDE-DE.png}
%         \caption{Execution Stage} % 這裡寫 (b) 的標題
%         \label{fig:ctde_exec}             % 這裡設 (b) 的標籤
%     \end{subfigure}

%     % --- 整個大圖的標題與標籤 ---
%     \caption{CTDE Paradigm} % 這是 Figure 5 的總標題
%     \label{fig:ctde_main}   % 這是 Figure 5 的總標籤
%     \vspace{-1em}
% \end{figure}

\begin{figure}[h] % [h] 代表盡量放在當前位置 (here)
    \centering % 讓圖片置中
    \vspace{-1em}
    \includegraphics[width=0.65\textwidth]{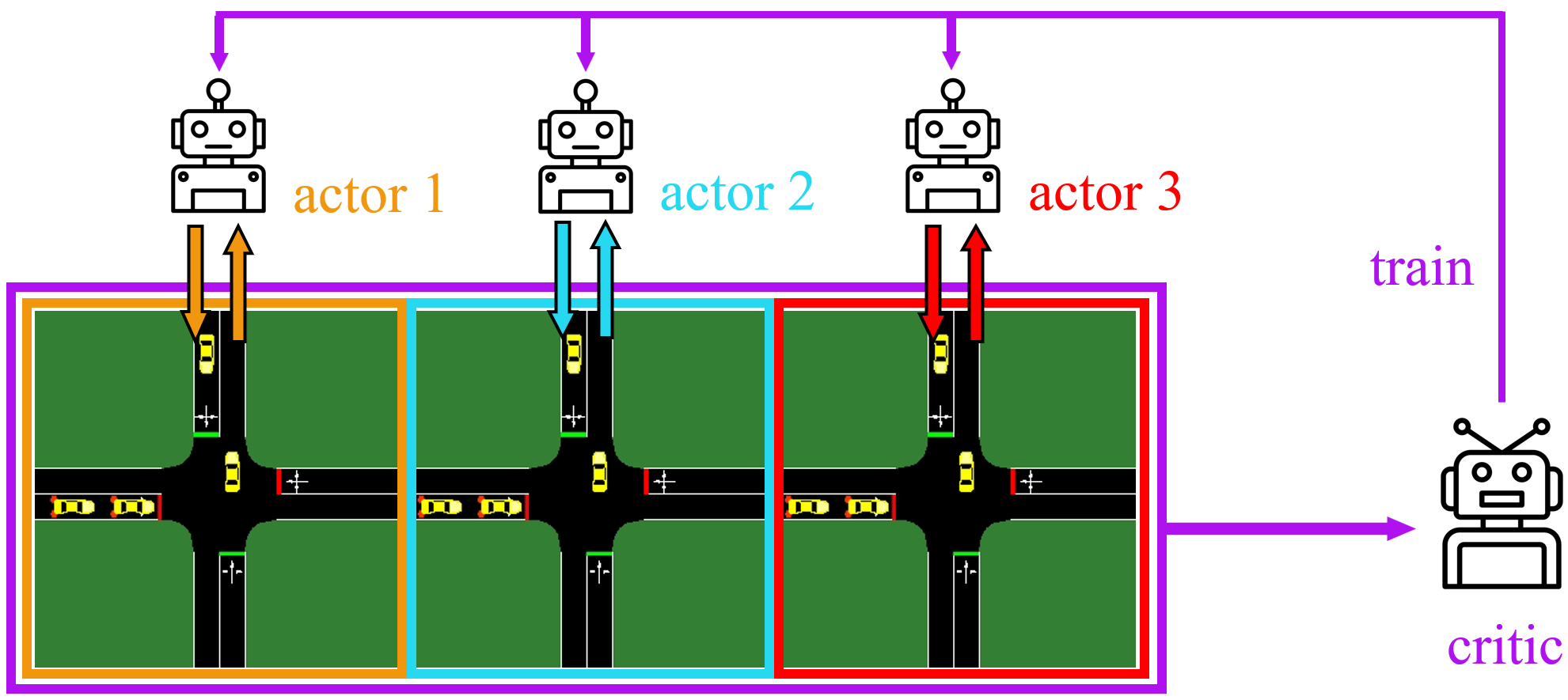} % 設定寬度為頁面文字寬度的 80%
    \caption{CTDE Paradigm} % 圖片標題
    \label{fig: CTDE} % 用於參照的標籤
    \vspace{-1em}
\end{figure}

We define the observation scope $\mathcal{O}^{\text{neighbor}}$ for each agent to include its local observation and the aggregated information from directly connected upstream and downstream neighbors $\mathcal{N}_i$:
\vspace{-0.3em}
\begin{equation}
    o_{i, t}^{\text{neighbor}} = \{ o_{i, t}^{\text{local}} \} \cup \{o_{j, t}^{\text{local}} \mid j \in \mathcal{N}_i \}
    \vspace{-0.3em}
\end{equation}
To implement this framework, we adopt the Multi-Agent Proximal Policy Optimization (MAPPO) algorithm \cite{yu_surprising_2022}.
It combines the stability of PPO's trust-region updates with the CTDE framework.
This mechanism enables agents to internalize cooperative behaviors and anticipate network-wide impacts, effectively achieving global coordination performance within a scalable, decentralized architecture.

\section{Experiments}
This section empirically validates the effectiveness and stability of the proposed MARL framework. 
We detail the experimental setup, present quantitative comparisons and ablation studies, and conclude with a qualitative stability analysis.

\subsection{Experimental Settings}
% 4.1 V1
% To ensure the experimental results hold significant practical value, we utilize PTV Vissim, one of microscopic traffic simulator. 
% Unlike simplified kinematic models (e.g., SUMO or CityFlow), Vissim employs the psycho-physical Wiedemann car-following model, which accurately captures the stochasticity and dynamic behaviors of human drivers, providing a high-fidelity testbed for Sim-to-Real validation.

\begin{figure}[h] % [h] 代表盡量放在當前位置 (here)
    \centering % 讓圖片置中
    \vspace{-0.5em}
    % --- 第一個子圖 (a) ---
    \begin{subfigure}[b]{0.45\textwidth}
        \centering
        % 這裡放左邊那張圖的檔名
        \includegraphics[width=\textwidth]{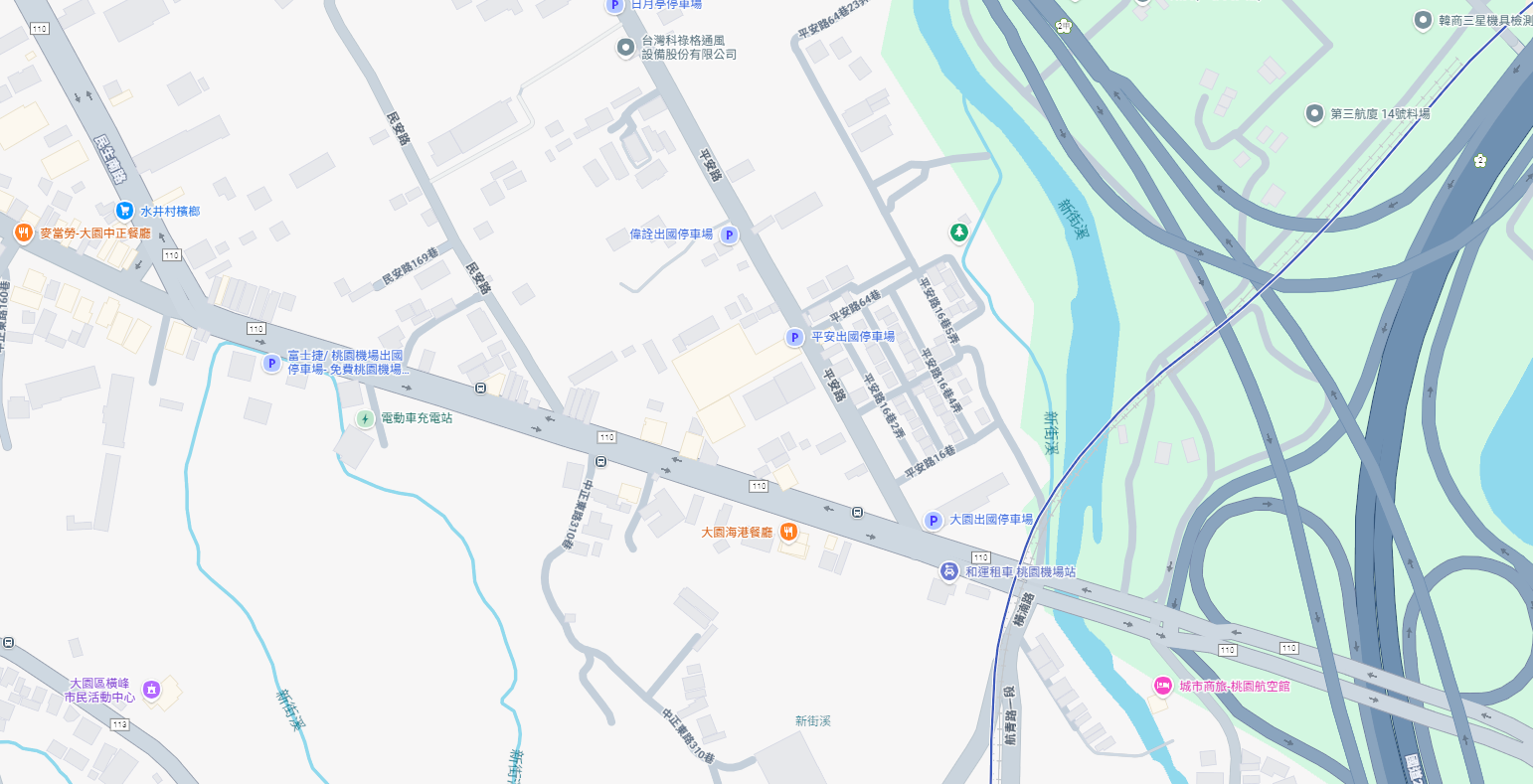} 
        \caption{Satellite Image} % 這裡寫 (a) 的標題
    \end{subfigure}
    \hfill % 這指令是用來把兩張圖撐開，讓它們並排
    % --- 第二個子圖 (b) ---
    \begin{subfigure}[b]{0.48\textwidth}
        \centering
        % 這裡放右邊那張圖的檔名
        \includegraphics[width=\textwidth]{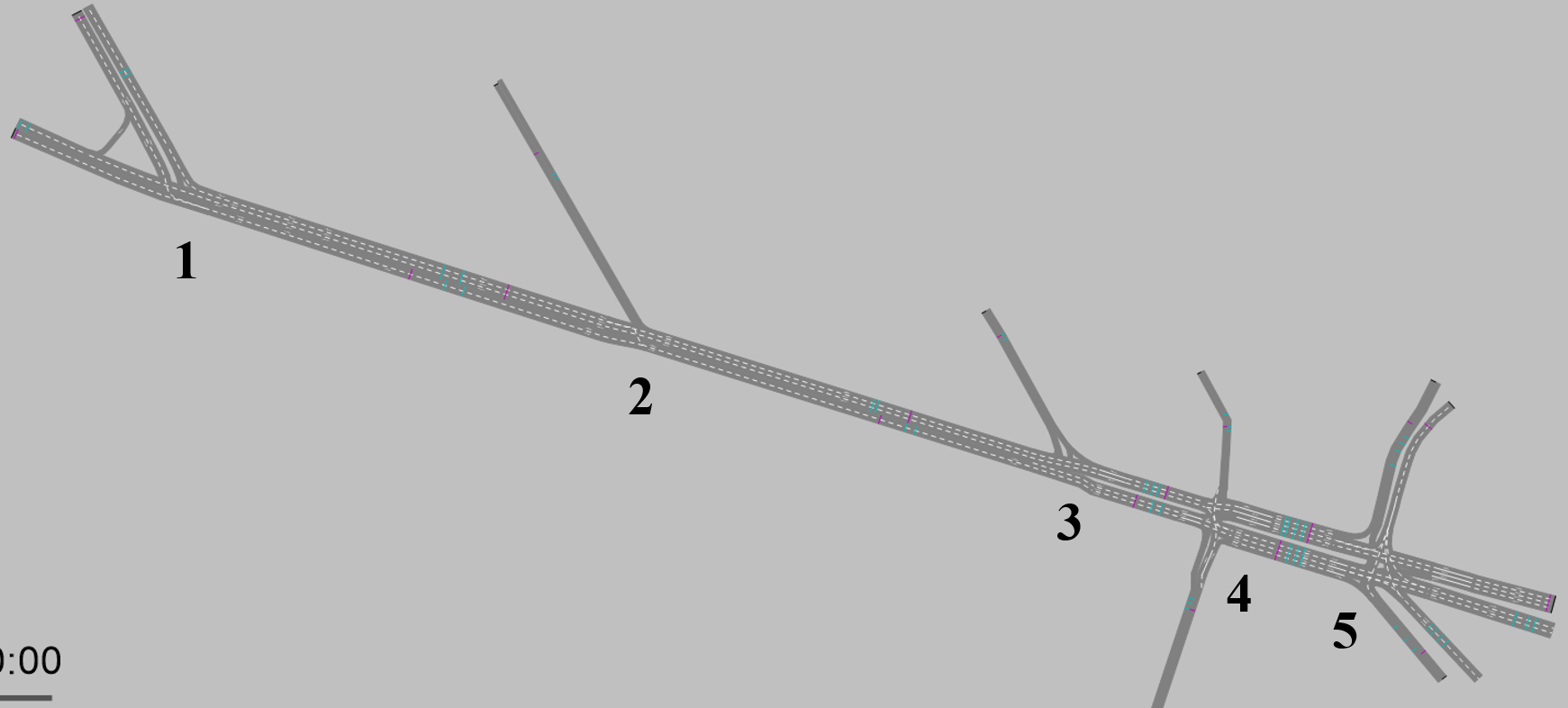}
        \caption{Vissim Modeling} % 這裡寫 (b) 的標題
    \end{subfigure}

    % --- 整個大圖的標題與標籤 ---
    \vspace{-0.5em}
    \caption{Experimental Network} 
    \label{fig:dayuan map}   
    \vspace{-0.5em}
\end{figure}

\textbf{Simulation Environment and Road Network}
To ensure practical relevance and fidelity, we utilize PTV Vissim, a microscopic simulator employing the psycho-physical Wiedemann car-following model \cite{wiedemann_simulation_1974}. 
Vissim accurately captures the stochastic dynamics of human driving behavior, serving as a robust testbed for sim-to-real validation.
The simulation environment is a calibrated digital twin of Zhongzheng East Road in Taoyuan City, Taiwan, comprising five consecutive signalized intersections characterized by short spacing and high interaction (\Cref{fig:dayuan map}). 
Network geometry, including lane configurations and turning pockets, was rigorously aligned with real-world satellite imagery and field surveys.

% 4.1 V1
% The experimental network is a calibrated digital twin of Zhongzheng East Road in Taoyuan City, Taiwan. 
% This arterial corridor consists of five consecutive signalized intersections characterized by short inter-intersection distances and high traffic interaction. 
% The network geometry, including lane configurations, pocket lengths for turning vehicles, and stop line positions, was rigorously calibrated against real-world satellite imagery and field surveys.

% 4.1 V1
% A common pitfall in RL-TSC research is the use of identical traffic distributions for both training and testing, which leads to biased performance evaluations. 
% To address this, we designed a Cross-Scenario Validation mechanism based on real-world data.

\begin{figure}[t]
    \centering
    
    % --- 左邊區塊：放圖片 (Figure 7) ---
    % [c] 代表垂直置中對齊，0.55\textwidth 代表佔據頁面 55% 寬度
    \begin{minipage}[c]{0.50\textwidth} 
        \centering
        % 插入你的圖片
        \includegraphics[width=\textwidth]{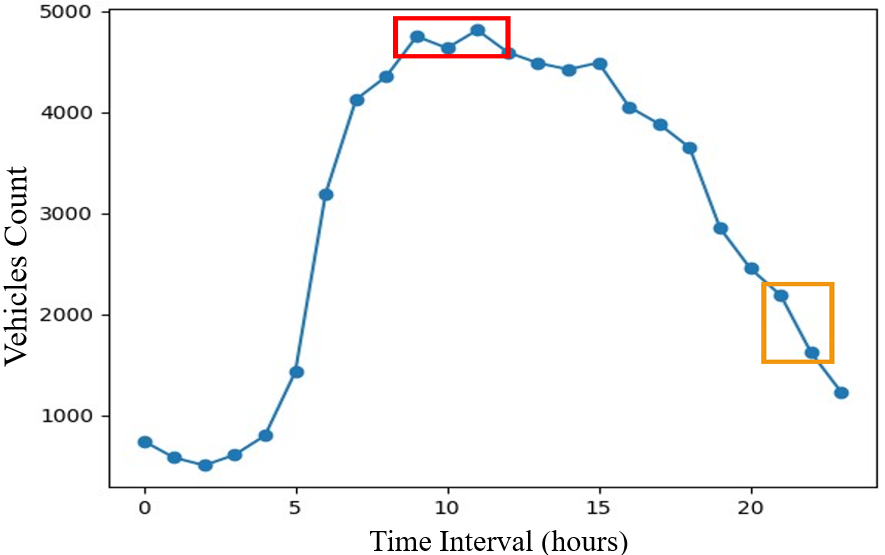} 
        % 圖片標題通常在下方
        % \vspace{-10pt}
        \caption{24-hour traffic volume}
        \label{fig:24-hour traffic volume}
        \begin{minipage}{0.85\linewidth}
        \vspace{0.5em} % 與標題留點距離
        \parbox{1\textwidth}{
            \textit{Note.} This 24-hour weekday count tracks vehicles entering the source link.
        }
        \end{minipage}
    \end{minipage}
    \hfill % 這指令很重要，用來把左右兩個區塊撐開
    % --- 右邊區塊：放表格 (Table 1) ---
    % 剩餘寬度給表格 (0.58 + 0.38 + 間隙 < 1.0)
    \begin{minipage}[c]{0.46\textwidth} 
        \centering
        % 表格標題通常在上方，這裡必須用 captionof
        \captionof{table}{Traffic volume} 
        \label{tab:traffic config}
        
        % 你的表格內容
        \resizebox{\textwidth}{!}{ % 如果表格太大，可以用這個縮放
            \begin{tabular}{lcc}
                \toprule
                \textbf{Flow} & \textbf{Data Time} & \textbf{\makecell{Vehicle Count\\(vehs/hr)}} \\
                \midrule
                Peak Hour & 9:00-10:00 & $\sim$ 4800 \\
                Off-Peak & 21:00-22:00 & $\sim$ 1800 \\
                \bottomrule
            \end{tabular}
        }
    \end{minipage}
    \vspace{-0.5em}
\end{figure}

\textbf{Traffic Data and Scenarios}
A critical challenge in RL research is the bias introduced by using identical traffic distributions for training and testing. 
To mitigate this, we implemented a cross-scenario validation mechanism. 
We analyzed 24-hour real-world detector data (\Cref{fig:24-hour traffic volume}) to extract two distinct stress levels: a high-load peak hour and a moderate-load off-peak hour (\Cref{tab:traffic config}). 
This paper only considers four-wheeled vehicles.
Agents are trained strictly on peak hour data to enforce policy optimization under high-pressure conditions, while evaluation is conducted on both scenarios to test generalization.

% \begin{figure}[h] % [h] 代表盡量放在當前位置 (here)
%     \centering % 讓圖片置中
%     \vspace{-1em}
%     \includegraphics[width=0.5\textwidth]{img/vehicle_number_per_hour.png} % 設定寬度為頁面文字寬度的 80%
%     \caption{24-hour traffic volume profile} % 圖片標題
%     \label{fig:24-hour traffic volume} % 用於參照的標籤
%     \begin{minipage}{0.85\linewidth}
%     \footnotesize
%         \textit{Note.} This line chart shows the total number of vehicles per hour on this road network, reflecting the distribution of traffic flow throughout the day.
%     \end{minipage}
%     \vspace{-1em}
% \end{figure}

% 4.1 V1
% Traffic volume data was collected from vehicle detectors installed along the corridor. 
% We analyzed the 24-hour flow profile and selected two distinct periods to represent different stress levels: Peak Hour and Off-Peak Hour. 
% Peak Hour represents a high-load, congestion-prone scenario, and Off-Peak Hour represents a moderate-load, free-flow scenario.

% \begin{table}[htbp]
%     \centering
%     \vspace{-1em}
%     % 表格標題 (Caption)
%     \caption{Traffic volume Settings}
%     \label{table:traffic config}
    
%     % 定義欄位：l=靠左, c=置中. 這裡設為: 左, 左, 中, 左
%     \begin{tabular}{llcl}
%         \toprule
%         % 表頭區
%         \textbf{Flow} & 
%         \textbf{Data Time} &
%         \textbf{\makecell[l]{Vehicle number\\(vehicles/hr)}}\\
%         \midrule
        
%         % 第一列資料 (Single-Intersection)
%         Peak Hour & 
%         9:00-10:00 &
%         \makecell[l]{$\sim 4800$} \\
        
%         \addlinespace % 增加一點行距，讓兩筆資料不要黏太緊
        
%         % 第二列資料 (Arterial-3)
%         Off-Peak Hour & 
%         21:00-22:00 &
%         \makecell[l]{$\sim 1800$} \\
        
%         \bottomrule
%     \end{tabular}
% \end{table}
% 4.1 V1
% We strictly use the Peak Hour data for training, and the trained models are evaluated on both the Peak Hour and the Off-Peak Hour.. 
% This forces the agent to learn policy optimization under high-pressure conditions, dealing with potential queue spillover and saturation.

\textbf{Comparative Methods}
% 4.1 V1
% To comprehensively evaluate the proposed framework, we conducted comparative experiments involving two established baselines and comprehensive analysis of our proposed methods.
% \textbf{Baselines:}
% \begin{itemize}[topsep=-5pt, nosep]
%     \item Fixed-Time (Real-World Plan): 
%         The actual signal timing plan currently deployed by the local transportation authority, optimized for arterial progression (Green Wave).
%     \item MaxPressure \cite{varaiya_max_2013}: 
%         A classic heuristic control algorithm that minimizes the pressure difference between upstream and downstream links.
% \end{itemize}
For comparative analysis, we benchmark our framework against a fixed-time plan optimized for green waves and the MaxPressure heuristic \cite{varaiya_max_2013}.
We evaluate our proposed methods using a unified notation $M_{scope}^{strategy}$, where the subscript denotes the observation scope (local, neighbor, global) and the superscript indicates the training strategy (static, randomized).

% 4.1 V1
% \textbf{Proposed Methods:} 
% we conduct a comprehensive analysis of the proposed framework across two key dimensions: Observation Scope and Training Strategy. 
% This full-factorial design allows us to isolate the contributions of each component.
% To facilitate the subsequent discussion and visualization of results, we introduce a unified notation system $\text{M}_{\text{scope}}^{\text{strategy}}$ to represent the different model variants:
% \begin{itemize}[topsep=-5pt, nosep]
%     \item Subscript (Scope): 
%         Denotes the observation scope of the agent. 
%         $\text{local}$, $\text{neighbor}$, and $\text{global}$ respectively represent local, neighbor, and global observation defined in \cref{observation scopes definition}.
%     \item Superscript (Strategy): 
%         Denotes the training strategy regarding turning ratios. 
%         $s$ (standard) represents the model trained using the static turning ratio (baseline RL approach).
%         $r$ (robust) represents the model trained with the proposed Turning Ratio Randomization.
% \end{itemize}

\textbf{Evaluation Metrics}
Following established evaluation protocols in recent TSC review \cite{noaeen_reinforcement_2022}, we quantify performance using four metrics: Average Travel Time (ATT, s/veh), Average Waiting Time (AWT, s/veh), Average Delay (AD, s/veh), and Vehicle Count (VC, vehs/h). 
% This selection ensures a multidimensional assessment covering both system-level throughput and individual user experience.

% 4.1 V1
% We employ four key metrics to quantify traffic efficiency and control performance:
% \begin{enumerate}[topsep=-5pt, nosep]
%     \item Average Travel Time (ATT, seconds/veh): 
%         The average duration required for a vehicle to traverse the network from entry to exit. 
%         $$\text{ATT} = \frac{1}{N_{veh}} \sum_{i=1}^{N_{veh}} (t_{\text{exit}, i} - t_{\text{enter}, i})$$
%     \item Average Waiting Time (AWT, seconds/veh):
%         The cumulative time all vehicles spend at a standstill (speed < 0.1 m/s) within the network.
%     \item Average Delay (AD, seconds/veh):
%         The difference between the actual travel time and the ideal free-flow travel time.
%         $$\text{AD} = \frac{1}{N_{veh}} \sum_{i=1}^{N_{veh}} (t_{\text{actual}, i} - t_{\text{ideal}, i})$$
%     \item Vehicles Per Hour (vph, veh/hr):
%         The total number of vehicles that successfully exit the network per hour. 
% \end{enumerate}

\subsection{Experimental Results and Analysis}
\Cref{tab:overall_comparison} presents the quantitative comparison of all methods across the four evaluation metrics under both peak and off-peak scenarios. 
The results reveal distinct performance patterns regarding robustness and generalization.

\begin{table*}[t]
    \centering
    \fontsize{11}{13}\selectfont
    \setlength{\tabcolsep}{3pt}
    \caption{Performance Comparison between Peak and Off-Peak Scenarios}
    \label{tab:overall_comparison}
    
    \resizebox{\textwidth}{!}{
        % Preamble: 定義欄位，同樣在中間加入 1cm 的強制留白
        \begin{tabular}{ll cccc @{\hskip 1cm} cccc}
            \toprule
            
            \multirow{2.5}{*}{\textbf{Category}} & \multirow{2.5}{*}{\textbf{Method}} & 
            
            \multicolumn{4}{c@{\hskip 0.8cm}}{\makebox[0pt]{\textbf{Peak Hour Flow}}} & 
            \multicolumn{4}{c}{\makebox[0pt]{\textbf{Off-Peak Hour Flow}}} \\
            
            % 線條修剪：左邊修右端(r)，右邊修左端(l)，讓視覺分離更明顯
            \cmidrule(r{2em}){3-6} \cmidrule(){7-10}
            
             & & \textbf{ATT} $\downarrow$ & \textbf{AWT} $\downarrow$ & \textbf{AD} $\downarrow$ & \textbf{VC} $\uparrow$ & \textbf{ATT} $\downarrow$ & \textbf{AWT} $\downarrow$ & \textbf{AD} $\downarrow$ & \textbf{VC} $\uparrow$ \\
            \midrule
            
            % --- Group 1: Baseline ---
            \multirow{2}{*}{Baseline} 
             & FixTime     & 383.92 & 352.87 & 319.04 & 4015.87 & 129.20 & 50.74 & 58.60 & 1789.93 \\
             & MaxPressure & 265.79 & 285.93 & 196.54 & 4223.80 & 126.57 & 45.96 & 55.82 & 1797.53 \\
            \midrule
            
            % --- Group 2: Standard RL ---
            \multirow{3}{*}{Standard RL} 
             & $M^{\text{static}}_{\text{local}}$    & 266.07 & 227.79 & 197.74 & 4446.93 & 132.20 & 51.42 & 61.32 & 1799.80 \\
             & $M^{\text{static}}_{\text{neighbor}}$ & 249.54 & \textbf{215.47} & 181.08 & \textbf{4448.13} & 130.27 & 50.02 & 59.59 & 1798.53 \\
             & $M^{\text{static}}_{\text{global}}$   & 262.24 & 230.62 & 194.05 & 4404.20 & 135.71 & 51.46 & 65.05 & 1787.20 \\
            \midrule
            
            % --- Group 3: Robust RL (Ours) ---
            \multirow{3}{*}{Robust RL (ours)} 
             & $M^{\text{randomized}}_{\text{local}}$    & 242.96 & 228.36 & 174.44 & 4316.80 & 129.37 & 48.05 & 58.47 & 1801.47 \\
             & $M^{\text{randomized}}_{\text{neighbor}}$ & \textbf{230.58} & 231.01 & \textbf{160.34} & 4416.53 & 124.37 & 44.09 & 53.44 & \textbf{1808.47} \\
             & $M^{\text{randomized}}_{\text{global}}$   & 256.39 & 219.30 & 188.08 & 4398.80 & \textbf{119.32} & \textbf{36.12} & \textbf{48.33} & 1802.80 \\
            
            \bottomrule
        \end{tabular}
    }
    \vspace{-1em}
\end{table*}

In the peak hour scenario (\Cref{tab:overall_comparison} left), which aligns with the training distribution, the proposed MARL framework demonstrates superior control efficiency compared to the baselines. 
Specifically, the robustly trained $M_{\text{neighbor}}^{\text{randomized}}$ achieves an ATT of 230.58s, significantly surpassing the competitive MaxPressure heuristic (265.79s). 
This confirms that our framework effectively learns efficient signal timing policies under high-congestion conditions.

The off-peak scenario (\Cref{tab:overall_comparison} right) serves as a critical test for generalization and highlights the impact of the training strategy. 
Results also reveal that standard RL models trained with static ratios fail to adapt to this unseen environment. 
Due to severe overfitting to specific training patterns, their performance degrades significantly, falling even behind the heuristic baselines. 
% ting: Capitalization. This is the name of a technique so maybe it's necessary. Let's check for consistency (seems consistent right now)
In contrast, Turning Ratio Randomization significantly enhances robustness. 
While the global-view agent $M_{\text{global}}^{\text{randomized}}$ naturally achieves the best performance (119.32s ATT), the neighbor-based agent $M_{\text{neighbor}}^{\text{randomized}}$ also performs remarkably well (124.37s ATT). 
It not only outperforms the MaxPressure baseline (126.57s ATT) but also approaches the efficacy of the global agent. 
% This indicates that neighbor-level observations, when processed via the CTDE framework, capture the essential spatial dependencies required to closely approximate global-level coordination.

Ultimately, this underscores the pivotal role of Turning Ratio Randomization as an essential regularization technique, equipping the agent with the necessary robustness to function reliably in dynamic, unseen environments where standard training falls short.

\subsection{Component Analysis}
To provide a deeper analysis of our framework's internal mechanisms, this section conducts targeted ablation studies on critical design choices.

First, we assess the necessity of the CTDE paradigm by benchmarking MAPPO against the non-CTDE algorithm IPPO. 
For a fair comparison, both algorithms utilize identical Neighbor-Level Observations and Turning Ratio Randomization, differing solely in their critic mechanism: IPPO uses a decentralized critic, whereas MAPPO employs the centralized critic.

\begin{table}[t] % 如果是雙欄論文，用 table* 可以跨兩欄；單欄論文用 table 即可
    \centering
    \fontsize{11}{13}\selectfont
    \setlength{\tabcolsep}{3pt}
    \caption{non-CTDE Algorithm v.s. CTDE Algorithm}
    \label{tab: ippo}
    
    % 使用 resizebox 確保表格剛好填滿頁面寬度 (可選)
    % \resizebox{\textwidth}{!}{
        \begin{tabular}{ll cccc @{\hskip 0.8cm} cccc}
            \toprule
            \multirow{2.5}{*}{\textbf{Category}} & \multirow{2.5}{*}{\textbf{Algorithm}} & 
            \multicolumn{4}{c@{\hskip 1cm}}{\textbf{Peak Hour Flow}} & 
            \multicolumn{4}{c}{\textbf{Off-Peak Hour Flow}} \\
            
            % cmidrule 用來畫中間的分組線，(lr) 可以讓線條左右內縮，比較好看
            \cmidrule(lr{2em}){3-6} \cmidrule(){7-10}
            
             & & \textbf{ATT} $\downarrow$ & \textbf{AWT} $\downarrow$ & \textbf{AD} $\downarrow$ & \textbf{VC} $\uparrow$ & \textbf{ATT} $\downarrow$ & \textbf{AWT} $\downarrow$ & \textbf{AD} $\downarrow$ & \textbf{VC} $\uparrow$ \\
            \midrule
            
            \multirow{2}{*}{Baseline} 
             & FixTime     & 383.92 & 352.87 & 319.04 & 4015.87 & 129.20 & 50.74 & 58.60 & 1789.93 \\
             & MaxPressure & 265.79 & 285.93 & 196.54 & 4223.80 & 126.57 & 45.96 & 55.82 & 1797.53 \\
            \midrule
            
            \multirow{2}{*}{RL-based} 
             & non-CTDE  & 298.43 & 319.72 & 231.89 & 4124.13 & 134.20 & 52.08 & 63.40 & 1790.20 \\
             & CTDE (ours) & \textbf{230.58} & \textbf{231.01} & \textbf{160.34} & \textbf{4416.53} & \textbf{124.37} & \textbf{44.09} & \textbf{53.44} & \textbf{1808.47} \\
            
            \bottomrule
        \end{tabular}
    % }
\end{table}

As shown in \Cref{tab: ippo}, MAPPO significantly outperforms IPPO. 
While IPPO suffers from environment non-stationarity and unstable credit assignment due to simultaneous neighbor updates, MAPPO leverages its global critic to guide local actors toward cooperative behaviors, thereby justifying the adoption of the CTDE framework.

\begin{table}[t] % 如果是雙欄論文，用 table* 可以跨兩欄；單欄論文用 table 即可
    \centering
    \fontsize{11}{13}\selectfont
    \setlength{\tabcolsep}{3pt}
    \caption{Action Space Adjustment Comparison}
    \label{tab: action space comparison}
    
    % 使用 resizebox 確保表格剛好填滿頁面寬度 (可選)
    \resizebox{\textwidth}{!}{
        \begin{tabular}{ll cccc @{\hskip 0.8cm} cccc}
            \toprule
            \multirow{2.5}{*}{\textbf{Category}} & \multirow{2.5}{*}{\textbf{Adjustments}} & 
            \multicolumn{4}{c@{\hskip 1cm}}{\textbf{Peak Hour Flow}} & 
            \multicolumn{4}{c}{\textbf{Off-Peak Hour Flow}} \\
            
            % cmidrule 用來畫中間的分組線，(lr) 可以讓線條左右內縮，比較好看
            \cmidrule(lr{2em}){3-6} \cmidrule(){7-10}
            
             & & \textbf{ATT} $\downarrow$ & \textbf{AWT} $\downarrow$ & \textbf{AD} $\downarrow$ & \textbf{VC} $\uparrow$ & \textbf{ATT} $\downarrow$ & \textbf{AWT} $\downarrow$ & \textbf{AD} $\downarrow$ & \textbf{VC} $\uparrow$ \\
            \midrule
            
            \multirow{2}{*}{Baseline} 
             & FixTime     & 383.92 & 352.87 & 319.04 & 4015.87 & 129.20 & 50.74 & 58.60 & 1789.93 \\
             & MaxPressure & 265.79 & 285.93 & 196.54 & 4223.80 & 126.57 & 45.96 & 55.82 & 1797.53 \\
            \midrule
            
            \multirow{2}{*}{Linear} 
             & Small-Scale & 263.11 & 289.70 & 196.79 & 4226.00 & 158.10 & 73.18 & 87.64 & 1775.87 \\
             & Large-Scale & 283.56 & 267.24 & 216.80 & 4229.60 & 144.96 & 59.32 & 74.09 & 1793.33 \\
            \midrule

             \multirow{2}{*}{Exponential} 
             & Base-2 (ours)  & \textbf{230.58} & 231.01 & \textbf{160.34} & 4416.53 & \textbf{124.37} & 44.09 & \textbf{53.44} & \textbf{1808.47} \\
             & Base-3 (ours) & 234.36 & \textbf{215.12} & 165.06 & \textbf{4445.80} & 125.98 & \textbf{43.11} & 54.86 & 1801.40 \\
            
            \bottomrule
            % \vspace{-pt}
            \multicolumn{10}{@{}p{15cm}@{}}{ % 10改成您的總欄位數
                \vspace{1pt} % 與上方分隔線留點距離
                \fontsize{11}{13}\selectfont
                \textit{Note:} Linear: Small-Scale \{0, $\pm 2$, $\pm 4$, $\pm 6$, $\pm 8$\}, Large-Scale \{0, $\pm 5$, $\pm 10$, $\pm 15$, $\pm 20$\}; Exponential: Base-2 \{0, $\pm 1$, $\pm 2$, $\pm 4$, $\pm 8$\}, Base-3 \{0, $\pm 1$, $\pm 3$, $\pm 9$, $\pm 27$\}
            }\\
        \end{tabular}
    }
    \par
    \vspace{-0.5em}
    
\end{table}
% ting: "Linear Adjustment" same problem with capitalization
Next, we evaluate the efficacy of the proposed action space design by comparing it against Linear Adjustment schemes. 
We tested two variants of Linear Adjustment: a standard setting $\{0, \pm2, \pm4, \pm6, \pm8\}$ and an extended linear set $\{0, \pm5, \pm10, \pm15, \pm20\}$ matched to the same action space size (9 actions) as our method.
Results (\Cref{tab: action space comparison}) indicate that our Exponential Adjustment yields superior performance across all metrics. 
Although Linear Adjustments can outperform the Fixed-Time plan in peak hours, they degrade significantly in off-peak scenarios. 
In contrast, the exponential distribution concentrates fine-grained actions near zero to minimize oscillation during stable flows, while simultaneously retaining large-magnitude actions to rapidly dissipate sudden queues. 
Consequently, the exponential design effectively balances the need for large-scale adjustments during congestion with the precision required for stable traffic.

\begin{figure}[h] % [h] 代表盡量放在當前位置 (here)
    \centering % 讓圖片置中
    \includegraphics[width=0.85\textwidth]{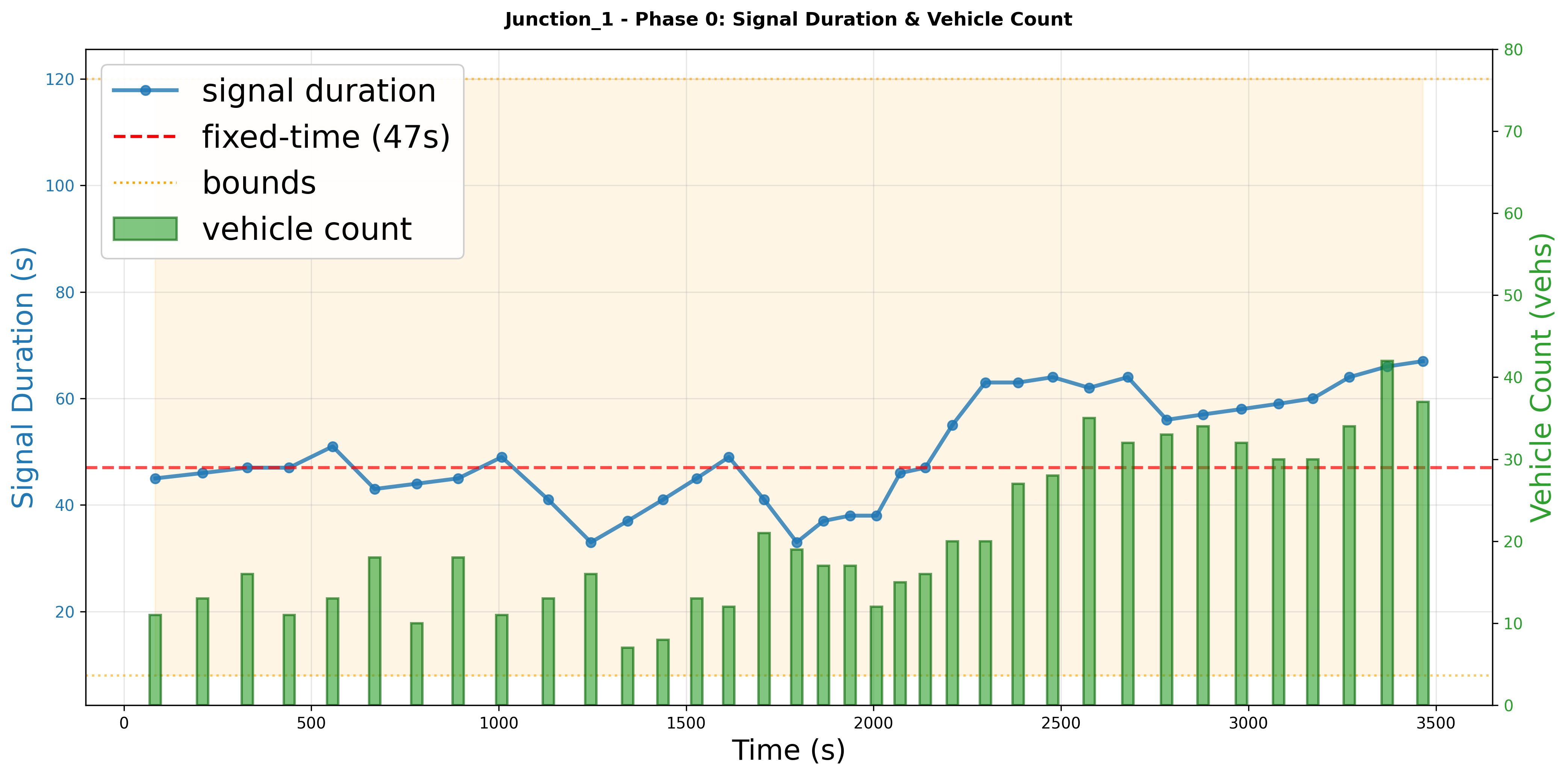} % 設定寬度為頁面文字寬度的 80%
    \vspace{-10pt}
    \caption{Synchronization of Signal Supply and Traffic Demand.} % 圖片標題
    \footnotesize
        \textit{Note:} The blue line and green bars track RL signal duration and traffic demand, respectively,  for the primary phase in the busiest intersection. 
        The red dashed line marks the fixed duration baseline of 47s. All durations fall within the yellow feasible action space.
    \label{fig: duration vs vehicle count}
    \vspace{-1em}
\end{figure}
% 4.3 V1
% Finally, to confirm the model's responsiveness to real-world dynamics, we analyze the temporal correlation between signal duration and traffic volume. 
% We select the 
% We recorded the vehicle count on the critical lane on main links to examine its relationship with the assigned green phase.
% As illustrated in \Cref{fig: duration vs vehicle count}, our method demonstrates exceptional adaptability. 
% The agent effectively scales the duration time in response to real-time demand fluctuations—conserving time during low demand and extending it during congestion—confirming its capability to handle dynamic traffic patterns.
% 4.3 V2
Finally, to validate the model's responsiveness to traffic dynamics, we analyze the temporal correlation between signal duration and traffic volume. 
We specifically selected the intersection with the highest traffic volume, comparing the green time allocated to its primary phase against the vehicle count on the corresponding critical lane on main links. 
As shown in Figure 8, our method exhibits strong adaptability. 
The agent dynamically scales signal duration to match real-time demand—reducing green time during low-traffic periods and extending it during congestion—thereby confirming its capability to handle dynamic traffic patterns.

\section{Conclusion}
% 5 V1
% This paper addresses the critical challenges of robustness, stability, and scalability in applying DRL to real-world Traffic Signal Control. 
% By leveraging a high-fidelity Vissim simulation environment, we proposed a comprehensive MARL framework incorporating three key technical innovations: Turning Ratio Randomization, Exponential Phase Duration Adjustment, and Neighbor-Based CTDE.

% Experiments demonstrated that training under static traffic conditions leads to severe overfitting. 
% In contrast, our Turning Ratio Randomization strategy effectively forces the agent to shift from "time-based memorization" to "state-based reactive control."
% Furthermore, by employing the MAPPO algorithm within a CTDE framework, agents using only Neighbor-Level observations achieved performance remarkably comparable to that of a centralized agent with global observation. 
% This validates our method as a scalable and communication-efficient solution suitable for large-scale arterial networks.

% Ultimately, this study bridges the gap between simulation and reality by preventing overfitting and ensuring control stability. 
% The proposed framework offers a promising path for deploying autonomous traffic signal control in the real world. 
% Future work will focus on extending this framework to grid-based networks and incorporating multi-modal traffic data (e.g., pedestrians and transit priority) to further enhance urban mobility.

% 5 V2
This paper addresses the critical challenges of robustness, stability, and scalability in applying DRL to real-world Traffic Signal Control. 
We proposed a high-fidelity MARL framework integrating three key innovations: Turning Ratio Randomization, Exponential Phase Duration Adjustment, and Neighbor-Based CTDE. 
Experimental results confirm that our randomization strategy effectively prevents overfitting by enforcing state-based reactivity. 
Furthermore, the CTDE paradigm successfully resolves the scalability-optimality dilemma, enabling agents with limited neighbor observations to achieve global-level coordination. 
By bridging the sim-to-real gap, this work offers a viable path for deploying autonomous signal control. 
Future research will extend this framework to grid networks and incorporate multi-modal traffic data to further enhance urban mobility.

\section{Acknowledgement} 

This work was supported in part by the National Science and Technology Council (NSTC) of Taiwan under Grant, NSTC 114-2221-E-A49-005, NSTC 114-2221-E-A49-006, and in part by ELAN Microelectronics Corporation, Taiwan.

% --- References ---
\renewcommand\refname{REFERENCES}
\fontsize{12}{14}\selectfont

% \bibitem{ref}
% Smith, A and Jones, B, Article Title, Journal, Publisher, Location, Date, pp. 1-10
% \bibliographystyle{IEEEtran}
\bibliographystyle{myIEEEtran}
\bibliography{ref}

\begin{thebibliography}{10}
\providecommand{\url}[1]{#1}
\csname url@samestyle\endcsname
\providecommand{\newblock}{\relax}
\providecommand{\bibinfo}[2]{#2}
\providecommand{\BIBentrySTDinterwordspacing}{\spaceskip=0pt\relax}
\providecommand{\BIBentryALTinterwordstretchfactor}{4}
\providecommand{\BIBentryALTinterwordspacing}{\spaceskip=\fontdimen2\font plus
\BIBentryALTinterwordstretchfactor\fontdimen3\font minus \fontdimen4\font\relax}
\providecommand{\BIBforeignlanguage}[2]{{%
\expandafter\ifx\csname l@#1\endcsname\relax
\typeout{** WARNING: IEEEtran.bst: No hyphenation pattern has been}%
\typeout{** loaded for the language `#1'. Using the pattern for}%
\typeout{** the default language instead.}%
\else
\language=\csname l@#1\endcsname
\fi
#2}}
\providecommand{\BIBdecl}{\relax}
\BIBdecl

\bibitem{inrix_scorecard_}
``{Global Traffic Scorecard}-{INRIX Global Traffic Ranking},'' Available at \url{https://inrix.com/scorecard/}.

\bibitem{krajzewicz_sumo_2002}
Krajzewicz, D., Hertkorn, G., Feld, C., and Wagner, P., \emph{{{SUMO}} ({{Simulation}} of {{Urban MObility}}); {{An}} Open-Source Traffic Simulation}, Jan. 2002.

\bibitem{zhang_cityflow_2019}
Zhang, H. \emph{et~al.}, ``{{CityFlow}}: {{A Multi-Agent Reinforcement Learning Environment}} for {{Large Scale City Traffic Scenario}},'' in \emph{The {{World Wide Web Conference}}}, May 2019, pp. 3620--3624.

\bibitem{_traffic_}
``Traffic {{Simulation Software}} \textbar{} {{PTV Vissim}} \textbar{} {{PTV Group}},'' Available at \url{https://www.ptvgroup.com/en/products/ptv-vissim}.

\bibitem{Chang_VissimRL_IV}
Chang, H.-C., Huang, S.-Y., Chen, Y.-C., and Wu, I.-C., ``{VissimRL}: {A Multi-Agent Reinforcement Learning Framework} for {Traffic Signal Control Based} on {Vissim},'' in \emph{IEEE Intelligent Vehicles Symposium (IV), arXiv:2601.18284}, Jun. 2026.

\bibitem{zang_metalight_2020}
Zang, X., Yao, H., Zheng, G., Xu, N., Xu, K., and Li, Z., ``{{MetaLight}}: {{Value-Based Meta-Reinforcement Learning}} for {{Traffic Signal Control}},'' \emph{Proceedings of the AAAI Conference on Artificial Intelligence}, vol.~34, no.~01, pp. 1153--1160, Apr. 2020.

\bibitem{ye_fairlight_2023}
Ye, Y., Ding, J., Wang, T., Zhou, J., Wei, X., and Chen, M., ``{{FairLight}}: {{Fairness-Aware Autonomous Traffic Signal Control With Hierarchical Action Space}},'' \emph{IEEE Transactions on Computer-Aided Design of Integrated Circuits and Systems}, vol.~42, no.~8, pp. 2434--2446, Aug. 2023.

\bibitem{wei_presslight_2019}
Wei, H. \emph{et~al.}, ``{{PressLight}}: {{Learning Max Pressure Control}} to {{Coordinate Traffic Signals}} in {{Arterial Network}},'' in \emph{Proceedings of the 25th {{ACM SIGKDD International Conference}} on {{Knowledge Discovery}} \& {{Data Mining}}}, ser. {{KDD}} '19.\hskip 1em plus 0.5em minus 0.4em\relax New York, NY, USA: Association for Computing Machinery, Jul. 2019, pp. 1290--1298.

\bibitem{wei_colight_2019}
Wei, H. \emph{et~al.}, ``{{CoLight}}: {{Learning Network-level Cooperation}} for {{Traffic Signal Control}},'' in \emph{Proceedings of the 28th {{ACM International Conference}} on {{Information}} and {{Knowledge Management}}}, Nov. 2019, pp. 1913--1922.

\bibitem{oroojlooy_attendlight_2020}
Oroojlooy, A., Nazari, M., Hajinezhad, D., and Silva, J., ``{{AttendLight}}: {{Universal Attention-Based Reinforcement Learning Model}} for {{Traffic Signal Control}},'' Oct. 2020.

\bibitem{chen_thousand_2020}
Chen, C. \emph{et~al.}, ``Toward {{A Thousand Lights}}: {{Decentralized Deep Reinforcement Learning}} for {{Large-Scale Traffic Signal Control}},'' \emph{Proceedings of the AAAI Conference on Artificial Intelligence}, vol.~34, no.~04, pp. 3414--3421, Apr. 2020.

\bibitem{cabrejas-egea_reinforcement_2021}
{Cabrejas-Egea}, A., Zhang, R., and Walton, N., ``Reinforcement {{Learning}} for {{Traffic Signal Control}}: {{Comparison}} with {{Commercial Systems}},'' \emph{Transportation Research Procedia}, vol.~58, pp. 638--645, Jan. 2021.

\bibitem{kumarasamy_integration_2024}
Kumarasamy, V.~K. \emph{et~al.}, ``Integration of {{Decentralized Graph-Based Multi-Agent Reinforcement Learning}} with {{Digital Twin}} for {{Traffic Signal Optimization}},'' \emph{Symmetry}, vol.~16, no.~4, Apr. 2024.

\bibitem{zhou_drle_2021}
Zhou, P., Chen, X., Liu, Z., Braud, T., Hui, P., and Kangasharju, J., ``{{DRLE}}: {{Decentralized Reinforcement Learning}} at the {{Edge}} for {{Traffic Light Control}} in the {{IoV}},'' \emph{IEEE Transactions on Intelligent Transportation Systems}, vol.~22, no.~4, pp. 2262--2273, Apr. 2021.

\bibitem{zhang_phase_2024}
Zhang, Z. \emph{et~al.}, ``Phase {{Re-service}} in {{Reinforcement Learning Traffic Signal Control}},'' Aug. 2024.

\bibitem{park_integrated_2024}
Park, J., Zhang, G., Wang, C., Wang, H., and Jiang, Z.-P., ``Integrated {{Routing}} and {{Traffic Signal Control}} for {{CAVs}} via {{Reinforcement Learning Approach}},'' in \emph{2024 {{IEEE}} 27th {{International Conference}} on {{Intelligent Transportation Systems}} ({{ITSC}})}, Sep. 2024, pp. 558--563.

\bibitem{wang_traffic_2024}
Wang, M. \emph{et~al.}, ``Traffic {{Signal Cycle Control}} with {{Centralized Critic}} and {{Decentralized Actors}} under {{Varying Intervention Frequencies}},'' \emph{IEEE Transactions on Intelligent Transportation Systems}, vol.~25, no.~12, pp. 20\,085--20\,104, Dec. 2024.

\bibitem{shah_deep_2024}
Shah, K.~D., Patel, D.~K., Raval, M.~S., Zaveri, M., and Merchant, S., ``Deep {{RL-Based Smart Signaling Using Space-Time Vehicular Features Under C-V2X Scenario}},'' in \emph{2024 16th {{International Conference}} on {{COMmunication Systems}} \& {{NETworkS}} ({{COMSNETS}})}, Jan. 2024, pp. 240--245.

\bibitem{han_wavelearner_2022}
Han, T., Lyu, S., and Oguchi, T., ``{{WaveLearner}}: {{A Knowledge-Combined Reinforcement Learning}} to {{Understand Coordinated Traffic Signal Control}} along {{Urban Arteries}},'' in \emph{2022 {{IEEE}} 25th {{International Conference}} on {{Intelligent Transportation Systems}} ({{ITSC}})}.\hskip 1em plus 0.5em minus 0.4em\relax Macau, China: IEEE Press, Oct. 2022, pp. 1167--1174.

\bibitem{yu_surprising_2022}
Yu, C. \emph{et~al.}, ``The {{Surprising Effectiveness}} of {{PPO}} in {{Cooperative}}, {{Multi-Agent Games}},'' Nov. 2022.

\bibitem{wiedemann_simulation_1974}
Wiedemann, R., ``{{SIMULATION DES STRASSENVERKEHRSFLUSSES}}.'' 1974.

\bibitem{varaiya_max_2013}
Varaiya, P., ``Max pressure control of a network of signalized intersections,'' \emph{Transportation Research Part C: Emerging Technologies}, vol.~36, pp. 177--195, Nov. 2013.

\bibitem{noaeen_reinforcement_2022}
Noaeen, M. \emph{et~al.}, ``Reinforcement learning in urban network traffic signal control: {{A}} systematic literature review,'' \emph{Expert Systems with Applications}, vol. 199, p. 116830, Aug. 2022.

\end{thebibliography}
% \printbibliography[title=REFERENCES]

\end{document}